# Benchmarking Foundation Speech and Language Models for Alzheimer's Disease and Related Dementia Detection from Spontaneous Speech

Jingyu Li, MS[a1], Lingchao Mao, MS[a1], Hairong Wang, MS[a], Zhendong Wang[b], PhD, Xi Mao, PhD[c*], Xuelei Sherry Ni, PhD[d*]

a. H. Milton Stewart School of Industrial and Systems Engineering, Georgia Institute of Technology, Atlanta, GA, USA
b. Shandong Mental Health Center, Jinan, Shandong Province, China.
c. Department of Economics, Robert C. Vackar College of Business & Entrepreneurship, University of Texas Rio Grande Valley, Edinburg, TX, USA
d. School of Data Science and Analytics, College of Computing and Software Engineering, Kennesaw State University, Kennesaw, GA, USA

**[1] co-first authors**

*** to whom correspondence should be addressed:**

**c***

Telephone: *956-882-8863*
Email: *xi.mao@utrgv.edu*
Address:
*1 W University Blvd BMAIN 2.522,*
*Brownsville, TX 78520*
*USA*

**d***

Telephone: *404-578-2251*
Email: *sni@kennesaw.edu*
Address:
*680 Arntson Drive, Kennesaw State University*
*Room 3411C, MD 9044*
*Marietta, GA 30060*
*USA*

# Abstract

## Background

Alzheimer’s disease and related dementias (ADRD) are progressive neurodegenerative conditions where early detection is critical for timely intervention and care planning. Spontaneous speech contains rich acoustic and linguistic markers that can serve as non-invasive biomarkers for cognitive decline. Foundation models, pre-trained on large-scale audio or text data, generate high-dimensional embeddings that encode rich contextual and acoustic information.

## Methods

In this study, we used Pioneering Research for Early Prediction of Alzheimer's and Related Dementias EUREKA (PREPARE) Challenge dataset which consists of audio recordings from over 1,600 participants with three distinct categories of cognitive decline: healthy control (HC), mild cognitive impairment (MCI), and Alzheimer’s Disease (AD). We further excluded samples that are non-English, non-spontaneous speech, or of poor quality. Our final samples included 703 (59.13%) HC, 81 (6.81%) MCI, and 405 (34.06%) AD cases. We systematically benchmarked a range of open-source foundation speech and language models to classify cognitive status into three categories (HC, MCI, or AD).

## Results

Whisper-medium model achieved the highest performance among speech models at 0.731 accuracy and 0.802 Area Under the Curve (AUC), while BERT with pause annotation achieved the top accuracy of 0.662 and 0.744 AUC among language models. Overall, ADRD detection based on state-of-the-art automatic speech recognition (ASR) model-generated audio-embeddings outperformed other models, and the inclusion of non-semantic information such as pause patterns consistently improved classification performance of text-embedding based models.

## Conclusion

Our work presents a comprehensive benchmarking framework built on state-of-the-art foundation models and validated on a large, clinically relevant dataset. Acoustic-based approaches—particularly ASR-derived embeddings— present great potential for the development of a more scalable, non-invasive, and cost-effective early detection tool for ADRD.

## Keywords

Alzheimer’s disease, deep learning, foundation models, automatic speech recognition

# 1 Introduction

## 1.1 Alzheimer's Disease and Early Detection

Alzheimer's disease and related dementias (ADRD) are progressive brain disorders that gradually impair memory, thinking, and daily functioning. ADRD typically progresses through three stages: healthy control (HC) or preclinical, mild cognitive impairment (MCI), and formal Alzheimer's disease (AD) or dementia [1] (**Figure 1**). MCI, while not always a precursor to AD, is an important early indicator and target for intervention [2].

Because ADRD progresses slowly and irreversibly, and new FDA-approved drugs like lecanemab and donanemab are effective only in early stages, timely detection is essential to delay symptoms and expand treatment options [3, 4]. Although advances in brain imaging, blood biomarkers, and artificial intelligence have improved early detection tools, global healthcare systems remain under-equipped. For instance, limited access to PET/MRI and dementia specialists leads to projected diagnostic delays of up to 10 years in the UK [5]. A six-country study found that less than half of individuals who reported memory problems received a formal diagnosis, and few underwent advanced testing—even when seeing a specialist [6]. In the U.S., early detection for ADRD is also inadequate and often delayed. Only 8% of Medicare enrollees receive a diagnosis of MCI (**Figure 1**), and more than 99% of primary care providers and clinics are identifying fewer cases than expected. Even specialists like geriatricians missed over 80% of early cases [7]. Estimates suggest that 14% of MCI cases progress to dementia annually [8] and over 7 million Medicare beneficiaries may miss treatment opportunities [9]. Missed and late diagnoses often disproportionately affect underrepresented populations, such as racial and ethnic minorities, rural communities, and those with lower socioeconomic status. The worsening health disparities in Alzheimer's care and outcomes highlight the critical need for accessible and scalable diagnostic solutions to support early detection of ADRD [10, 11].

## 1.2 Language Analysis for ADRD Detection

To close the gap between need and access, healthcare systems are exploring tools that support earlier, faster, and more affordable ADRD detection. One promising direction involves analyzing paralinguistic and linguistic patterns of speech, since language is among the earliest and most visible domains affected by ADRD [12, 13].

Common language symptoms in AD patients include naming difficulties, repetition, vague word use, inappropriate pronouns, etc [14, 15]. These often lead to fluent yet uninformative speech, with reduced coherence and incomplete sentences [16]. Temporal speech changes –such as increased hesitations and slower tempo– also occur, reflecting cognitive challenges in planning and organizing language [17, 18].

As a result, linguistic and paralinguistic features have been widely studied as non-invasive biomarkers of ADRD, supported by natural language processing and machine learning [19-21]. Markers such as semantic fluency, speech recognition, and acoustic features have proven effective in distinguishing normal aging from early MCI [21-23].These tools could be embedded in telehealth workflows, including virtual memory testing and tele-neuropsychology, which can be used for remote screening and triaging of patients. These services have been well-received by patients and help reduce diagnostic delays [24, 25].

### 1.3 Modern Foundation Speech and Language Models for ADRD Detection

Traditional machine learning approaches for ADRD detection have relied on manually engineered features, such as lexical diversity, syntactic complexity, coherence from transcripts, and acoustic features like pause duration, reaction time, and speech rate [19-21, 23, 26, 27]. These are then used in machine learning classifiers. However, these pipelines are labor-intensive and depend on controlled environments. Robin et al. [22] stressed the importance of rigorously evaluating recording quality and data processing accuracy.

Recent advances in deep learning have led to highly capable foundation speech and language models by pre-training deep learning architectures on large-scale datasets. These models learn to capture rich contextual information and produce high-dimensional representations of raw speech or texts –known as *embeddings*– thus simplifying the labor-intensive and human-dependent pre-processing step. These embeddings capture semantic, syntactic, and prosodic signals and have shown promise in ADRD detection [28-30]. Leveraging embeddings generated by large language and speech models also offers greater flexibility, robustness to varied inputs, and even multi-lingual capabilities. This shift supports more generalizable and scalable detection in telehealth or low-resource settings.

Two common approaches exist for extracting these embeddings for ADRD detection. One approach is to transcribe speech to text and then use a language model to extract text embeddings [28, 29]. However, the transcript-based approach has two major limitations. First, transcription errors and uncertainty might degrade model performance. Second, this method captures only semantic information, while neglecting important non-semantic information such as paralinguistic features and pauses patterns [29, 30]. To address these gaps, some researchers have manually annotated speech pauses in transcripts to preserve non-semantic information contained in patients' speech data [15, 29, 30]. This step has been shown to improve prediction accuracy, but it introduces additional complexity, lacks flexibility, and captures limited features characteristics of speech. Therefore, current research was limited to binary ADRD detection using small or curated datasets.

An increasingly promising alternative is to extract embeddings directly from raw audio using pre-trained automatic speech recognition (ASR) models trained for both recognition and representation tasks. These models preserve both semantic and paralinguistic cues without requiring transcription. Among them, OpenAI's Whisper has demonstrated state-of-the-art performance across varied languages and acoustic conditions [16]. However, despite their potential, these models have not yet been rigorously benchmarked for ADRD detection in large and real-world speech datasets.

In this paper, we address the aforementioned critical gaps by benchmarking a suite of modern automatic language models for ADRD detection, with a focus on real-world applicability, scalability, and cost-effectiveness. Specifically, we compare models based on text and audio embeddings generated by several state-of-the-art open-source foundation models, including language models such as BERT-family models for text embeddings and 15 speech models. We evaluate whether the high-dimensional embeddings generated by these models can serve as effective biomarkers for ADRD detection. To the best of our knowledge, neither the BERT-family models nor the speech models evaluated in this study have been systematically and comprehensively assessed for ADRD detection on a large-scale dataset. Our experiments utilize the Pioneering Research for Early Prediction of Alzheimer's and Related Dementias EUREKA (PREPARE) Challenge dataset—a large, diverse, and publicly available speech corpus curated from DementiaBank containing over 1,600 samples to ensure relevance to diverse population. Notably, our findings identify the medium-sized Whisper model as the top performer in ADRD detection from speech and demonstrate its advanced language understanding to capture linguistic and acoustic biomarkers associated with ADRD. We further discuss an acoustic-based ADRD detection pipeline with strong potential to advance early detection and intervention compared to traditional diagnosis timeline (**Figure 1)**. This study highlights the promise of foundation speech models for early ADRD detection and offers a practical guidance for the development of scalable and non-invasive ADRD screening tools. **Table 1** provides a summary of the significance of our work.

**Table 1.** Statement of Significance

| | |
|---|---|
| **Problem** | Current clinical assessments for ADRD are time-consuming, expensive, and delayed, limiting early diagnosis and effective treatment. |
| **What is Already Known** | Spontaneous speech contains linguistic and paralinguistic markers of cognitive decline, and prior work has explored deep learning models for binary ADRD detection using small or curated datasets. |
| **What this Paper Adds** | This study systematically benchmarks state-of-the-art open-source foundation speech and language models for ADRD detection on a large, clinically relevant dataset with multi-class classification (HC, MCI, AD). It demonstrates that audio embeddings from foundation speech models—especially Whisper—outperform text-based approaches and achieve promising detection accuracy. This work also discusses a practical acoustic-based pipeline for accessible and non-invasive ADRD screening. |
| **Who would benefit from the new knowledge in this paper** | Clinicians, researchers, and healthcare technologists developing early screening tools for ADRD, as well as public health stakeholders seeking accessible solutions for population-level dementia surveillance and intervention. |

# 2 Materials and Methods

## 2.1 The PREPARE Challenge Audio Data

The audio data used in this study is available as a part of the PREPAREPhase 2 Challenge – Acoustic Track-35 [31]. This dataset was curated from the DementiaBank [32-34], which contains multilingual audio recordings of participants performing a wide range of spontaneous speech tasks including picture description, verbal fluency, and sentence construction. The dataset has previously demonstrated its potential for early detection of ADRD during the PREPARE Phase 1 challenge. The original curated dataset consists of audio recordings from 2,086 participants who are given one of the following three cognitive status label: HC, MCI, or AD. These were further split into 1,646 training samples and 412 testing samples during the phase 2 challenge.

In this study, we used only the training dataset, as the cognitive status labels for the testing data were not released. The original dataset comprised speech samples from multiple languages and various language tasks, including sentence reading and spontaneous speech tasks. In the spontaneous speech task, participants were asked to describe the Cookies Theft picture for English speakers, which is a popular language impairment assessment tool originally used in the Boston Diagnostic Aphasia Examination protocol [17] Since speech and language models primarily analyze speech content, we focused our analysis exclusively on spontaneous speech samples. To identify spontaneous speech samples, we applied density-based spatial clustering of applications with noise (DBSCAN) on BERT embeddings of the speech transcriptions, followed by human evaluation to classify clusters to language tasks [18] [35]. See more details in **Figure S1**. The transcription and language of recordings were detected using the Whisper-small model [16]. Given that the majority of the recordings were in English, we excluded 319 non-English samples. Additionally, we removed 55 recordings of poor quality, defined as those with participant speech lasting less than three seconds or excessive background noise based on human review. Demographic information of each participant including age and gender is also available. A detailed description of data inclusion criteria is shown in **Figure 2**. The final dataset included recordings from 1,189 participants (≤30 seconds), including 703 classified as HC, 81 as MCI, and 405 as AD. The final sample (n=1,189) included 680 females and 509 males, with a mean (standard deviation) age of 75.178 (8.430).

## 2.2 Modeling Pipelines

We explore two machine learning pipelines to classify cognitive status of the participants from audio recordings: (i) a **text-based pipeline**, where a language model analyzes the transcribed content of the speech to identify linguistic patterns associated with cognitive decline; and (ii) an **audio-based pipeline**, where a speech model transforms audio signals into embeddings that capture both the linguistic and acoustic patterns characteristic of ADRD. **Figure 3** shows the detailed process of using both text- and audio-based pipelines for early detection of cognitive decline from spontaneous speech data.

Additionally, we compared the performance of the two deep learning-based pipelines with the traditional feature engineering pipeline, where classification models were trained on widely used pre-extracted paralinguistic and linguistic features reported in the literature including Geneva Minimalistic Acoustic Parameter Set (eGeMAPS) [36], ComParE [37], Mel-Frequency Cepstral Coefficients (MFCC) [38], and the Linguistic Inquiry and Word Count (LIWC) [39]. More details about this pipeline are provided in section **S 2**.

### 2.2.1 Text-based Pipeline

For the text-based pipeline, we first transcribed the preprocessed audio recordings into text using an ASR system, followed by the extraction of embeddings using pretrained language models for downstream classification. This approach emphasizes the linguistic content of speech for dementia diagnosis and can benefit from the abundance of textual data available on the internet, public datasets, and open-domain corpora. To incorporate acoustic characteristics into text embeddings, we further encoded pauses manually within the transcribed text to generate pause-aware embeddings.

**Audio preprocessing:** Each raw audio clip (48 kHz) was resampled to 16 kHz as higher frequencies in the audio recordings typically do not contain relevant speech. Audio amplitudes were normalized, and beginning and end silence were trimmed. Because recordings were collected in various acoustic environment, we used DeepFilterNet to suppress background noise and enhance audio clarity [40].

**Audio transcription**: We used OpenAI's Whisper model to transcribe audio recordings into text [16]. Preprocessed audio files were loaded and transformed into log-Mel spectrograms, which were fed into the Whisper model for transcription. Encoding pauses in the transcripts has shown superior predictive power in identifying AD status from audio data [29] To capture speech disfluencies relevant to cognitive decline, we included an optional step of annotating word pause and sentence pause in the transcripts. Specifically, word pauses were detected by measuring the time gap between consecutive words within a segment, and a pause marker was inserted if the duration exceeded a predefined threshold of 0.05 seconds. Sentence pauses were identified by calculating the silence duration between adjacent transcription segments, with pauses above the threshold labeled accordingly. After transcription, the text underwent further preprocessing, including formatting and tokenization, to ensure compatibility with text representation models. Specifically, we removed punctuation, adjusted text casing based on model requirements, and recoded word and sentence pauses for transcripts with the optional pause annotation. Similar to the approach introduced in Yuan et al.[29], we categorized pauses into three bins: short (under 0.5 sec); medium (0.5-2 sec); and long (over 2 sec), which were represented using punctuation marks—commas (,), periods (.), and ellipses (...), respectively to represent different levels of disfluencies in speech. The processed text underwent proper tokenization before served as input to text representation models.

**Text embedding models**: Bidirectional Encoder Representations from Transformers (BERT) [35] and BERT-like models have been highly effective for a wide range of natural language processing tasks, including AD classification. BERT is a transformer-based language model designed to learn contextualized word representations by leveraging bidirectional self-attention to capture associations among words. Each attention head processes elements within a sequence and generates a new sequence by computing a weighted sum of the transformed input representations. The base model consists of 12 layers, 12 attention heads, and 110M parameters, while the larger version has 24 layers, 16 attention heads, and 340M parameters. We also included two BERT models adapted for the medical domain in our benchmarking study. BioBERT [41] is a domain-specific adaptation of BERT, pretrained on large-scale biomedical corpora, including PubMed abstracts and full-text articles. We also evaluated BioClinicalBERT, a model further pretrained on clinical notes [42]. While BioBERT and BioClinicalBERT retain the transformer architecture and bidirectional pretraining approach of the base version of BERT, their exposure to domain-specific corpora allows the model to generate more relevant embeddings for medical and clinical applications. In our implementation, mean-pooled sentence-level embeddings were used for downstream classification.

### 2.2.2 Audio-based Pipeline

For the audio-based pipeline, we directly extracted speech embeddings from the recordings using foundation speech models designed from speech processing. This approach benefits from capturing phonetic, prosodic, and acoustic features relevant to speech impairments associated for dementia diagnosis.

**Audio preprocessing**: Each raw audio clip (48 kHz) was resampled to 16 kHz, matching the training input of most audio models, and audio amplitudes were normalized. Clips were then padded or truncated to an equal length of 30s segments. The preprocessed audio inputs were passed into model encoders to obtain mean pooled embeddings for each clip. For Whisper models, we converted each resampled clip into 80-channel (Whisper-medium or smaller) or 128-channel (Whisper-large or larger) log-Mel spectrogram representation on 25-millisecond windows with a stride of 10 milliseconds, following the original preprocessing of OpenAI [16].

**Audio embedding models**: We utilized a range of open-source ASR models and self-supervised speech models to learn meaningful speech representations from audio data. Whisper [16] is a state-of-the-art open-source transformer-based ASR model, trained on a large-scale multilingual dataset. It consists of an encoder-decoder architecture where the encoder processes raw audio into latent representations of audio features, and the decoder generates transcriptions. The family of Whisper models vary in size, with the larger versions offering improved transcription accuracy at the cost of higher computational demand. In our study, we demonstrated cases of using both encoder and decoder output where encoder output served as embedded features and decoder output served as audio transcripts for further processing.

We also compared a few prior open-source deep learning models for speech representation. Wav2vec2 [43] learns speech representations through a two-stage training process: a multi-layer convolutional neural network feature extractor encodes raw speech into latent representations, followed by a transformer encoder that refines these representations via masked speech prediction. HuBERT [44] utilizes a masked prediction approach where it learns to predict cluster-based speech units from unlabeled audio. This self-supervised learning method allows the model to capture speech structures without requiring transcriptions. Unispeech [45] improves upon prior models like Wav2Vec2 and HuBERT by integrating both unlabeled speech pretraining and supervised fine-tuning within a single model. WavLM [46] extends the Wav2Vec2 framework by incorporating denoising and masked speech modeling objectives in pre-training. Data2Vec further generalizes self-supervised learning by predicting latent representations instead of specific speech tokens, thus a more unified architecture across multiple modalities including speech [47].

### 2.2.3 Classifier and Covariates

Both pipelines were evaluated and compared on their ability to accurately classify AD. We mean-pooled sentence-level embeddings to form clip-level embeddings. We incorporated the patient's demographic information including age and sex by concatenating these two scalars to the final encoder embeddings of all the models. These modified embeddings then served as input features to a single-layer feedforward neural network which was trained to predict patient's cognitive status: HC, MCI or AD).

## 2.3 Experiment Setup

### 2.3.1 Model Architecture and Hyperparameters

For the text-based pipeline, we used Whisper-small and Whisper-medium for transcription. We did not consider larger Whisper models, as transcripts generated by Whisper-medium did not yield performance improvements over those by Whisper-small. We used pre-trained BERT ("bert-uncase-base") and BioBERT ("dmis-lab/biobert-v1.1") to extract embeddings from transcribed text. Each transcript was tokenized using the BERT's default tokenizer. The text was padded to a max length of 512 tokens or truncated if too long. The final embedded features extracted from both BERT and BioBERT models are of shape (768,).

For the audio-based pipeline, we evaluated 15 deep learning models designed for speech processing: Whisper (tiny, base, small, medium, large) [16], WavLM (base, base-plus, large) [46], Wav2Vec2 (base, large) [43], HuBERT (base, large) [44], Unispeech (large) [45], and Data2VecAudio (base, large) [47]. The extracted embeddings varied in dimensionality, with Whisper-tiny embeddings of shape (512,); Whisper-base, small, WavLM, Wav2Vec2, and HuBERT embeddings of shape (768,); Whisper-medium, Unispeech, and Data2VecAudio embeddings of shape (1024,); and Whisper-large embeddings of shape (1280,). The models were pretrained on different data corpora, from the 960-hour LibriSpeech dataset [48] used by (Wav2Vec2, HuBERT, and Data2Vec), to more diverse multisource pretraining corpus used by WavLM, and larger datasets like the 680,000-hour combination of public and proprietary speech data used by Whisper.

For both pipelines, we used a classification layer consisting of a single-layer feedforward neural network of size 128.

Cross Entropy loss was used for the multiclass classification. During training, the embedding extractors were kept frozen, and only the classification layer was trained.

Models were trained for up to 100 epochs with a batch size of 32, using the Adam optimizer with a learning rate of 0.0005 and ReduceLROnPlateau scheduler. To prevent overfitting, we terminated the training if the validation loss failed to improve for five consecutive epochs. All pre-trained models were sourced from the HuggingFace Transformers library and trained using PyTorch on an NVIDIA V100 GPU. The data preprocessing steps were performed using the pandas, numpy, pandas, audiofile, and librosa packages in Python. Evaluation metrics were computed using sklearn and torchmetrics packages in Python.

#### 2.3.2 Evaluation

The dataset was randomly divided into 80% training and 20% testing based on stratified sampling on the patient's labels of cognitive status to ensure balanced class proportions. 20% of the training set were used for validation. The train-test split was repeated five times with different random seeds. We reported the mean and standard deviation of accuracy and area under the curve (AUC) on the test set across five repetitions.

## 3 Results

**Table 2** shows the results of BERT, BioBERT and BioClinicalBERT using transcriptions generated by Whisper-small and Whisper-medium models. We reported average prediction accuracy and AUC across five repetitions. Among all configurations, BERT-base with pause annotations on Whisper-small transcriptions achieved the highest accuracy of 0.6622 ($\pm$0.0131), while the same model using Whisper-medium transcription achieved the highest AUC of 0.7444 ($\pm$0.0136). For the same type of model and configuration, incorporating pause annotations consistently led to improved performance—in terms of both accuracy and AUC—compared to their counterparts without pause annotations. we did not observe consistent performance gains from using a larger transcription model (Whisper-medium vs. Whisper-small).

**Table 3** presents the average performance on test set across five repetitions for each audio representation model. Overall, the family of Whisper-based models generally outperformed other audio representation models, likely due to their training on a substantially larger and more diverse multitask dataset. Specifically, the medium-size variant achieved the best performance with an accuracy of 0.7307 (±0.0202) and an AUC of 0.8024 (±0.0143). The three-class confusion matrix of the best performing model is presented in **Figure 4**. Models pretrained on a larger data corpus, such as Whisper, WavLM-base-plus, WavLM-large, generally outperformed those pretrained on smaller datasets, such as Data2Vec, Wav2Vec, and WavLM-base. While we noticed an increasing performance in both accuracy and AUC as we used a lager variant from the Whisper family from tiny to medium, Whisper-larger had worse performance than Whisper-medium. This pattern suggests that scaling up the model beyond a certain point does not necessarily translate into better classification performance. Fine-tuning the last layers of Whisper models did not improve performance (**Table S1**).

Comparing results of both text and audio representation models, we noticed that text-representation models, especially those with pause annotation, achieved comparable performance with audio representation models. We noticed a significant increase in both accuracy and AUC in the family of Whisper models.

**Table 4** summarizes the results of traditional feature engineering pipelines across five repetitions. Among all traditional features, the eGeMAPS achieved the highest accuracy of 0.643 (±0.020) and an AUC of 0.767 (±0.013)—performance comparable to several foundation speech models such as Unispeech, WavLM, Wav2Vec2, and Data2Vec.

# 4 Discussion

Our results show that non-semantic information is useful in dementia detection, similar to findings in previous literature [29, 30, 49-52]. In our text-based pipeline, we observed that language models tend to achieve higher accuracy and AUC when transcripts include annotated pauses. This suggests that non-semantic features, particularly pauses, may reflect patterns of cognitive decline that are useful for model training. Furthermore, state-of-the-art speech models consistently outperformed text-only models. This suggests that speech models are capable of retaining rich non-semantic information which is indicative of cognitive decline.

We noticed that using traditional acoustic features alone can achieve prediction accuracy comparable to several foundation models. However, traditional pipelines rely on manual pre-selection of acoustic and linguistic features, which may not fully capture the complex and subtle biomarkers of ADRD. In contrast, embedding-based foundation models, particularly those derived from large-scale pre-trained speech models, offer greater flexibility and data-driven representation learning. The Whisper-medium model significantly outperformed both the traditional feature-based models and other embedding models, highlighting its strong capacity to capture relevant speech characteristics for ADRD detection without the need for handcrafted features.

While our best model achieved an accuracy of 0.7307, this is lower than the 0.896 accuracy reported by Yuan et al.[29] We believe this difference may be due to the greater complexity of our classification task and the diversity of our dataset. Unlike previous work that focused on binary classification (HC vs. AD), we aimed to classify samples into three categories (HC, MCI, and AD). Moreover, our dataset includes nearly 1,200 samples with class imbalance, whereas the ADReSS challenge dataset used in most prior works is a curated dataset with around 200 samples balanced by age, gender, and diagnosis category [52]. While this added variability increases the difficulty of achieving high accuracy, the dataset used in this study reflects a larger and more diverse population, which enable a more realistic and comprehensive evaluation than those conducted in prior studies.

A machine learning–based tool for automatic AD screening via speech holds strong promise for providing an accessible and scalable workflow for early detection. Such a system can be deployed via a mobile application, allowing individuals to perform regular, low-effort speech-based screenings from their homes (**Figure 1**). Participants would be guided through simple spoken tasks—such as describing a picture or narrating a recent experience—with spontaneous speech captured directly through their device's microphone. The app would process the audio locally or via a secure cloud service using a pre-trained foundation speech model (e.g., Whisper) to assess cognitive risk. For healthy individuals, periodic screenings—such as once every three months—may suffice for proactive monitoring. If a user's speech is flagged as high risk for MCI, the app would recommend follow-up through an in-clinic cognitive assessment to confirm diagnosis. Early identification of MCI allows for timely intervention, personalized treatment planning, and patient education. Once diagnosed, individuals could continue using the app at higher frequency (e.g., weekly) to monitor their risk score progression and treatment response. This approach supports continuous, personalized cognitive health management in a non-invasive, scalable, and cost-effective manner. By supporting early, low-cost, and remote screening, foundation speech models offer a path toward more accessible Alzheimer's care and improved clinical outcomes across diverse real-world environments.

There are limitations of our study. First, our dataset may not fully represent the broader population in terms of language background, education level, or dialect variation which could affect acoustic characteristics and thus limit the generalizability of our findings. Additionally, while embeddings extracted from foundation models provide stronger performance compared to traditional acoustic features, they function as black boxes with limited interpretability. The lack of explainability may hinder clinical adoption and understanding of what specific features the models use to make predictions.

Future research could explore multi-lingual models to include non-English samples from the dataset such as Spanish, Gallego, and Chinese samples. Additionally, incorporating multimodal feature sets—such as participant's socioeconomics, medical history, and lifestyle information—could provide a more interpretable and comprehensive analysis of the individual's cognitive status and enhance model performance. As ADRD is a progressive disorder, it would be interesting to collect longitudinal speech data in the future and develop models that are effective at detecting subtle within-subject changes and tracking disease progression over time.

## 5 Conclusion

Our study presents a comprehensive benchmark for ADRD early detection across several foundation speech and language models and traditional machine learning models on a large, diverse, and clinically relevant dataset. Compared to pre-trained language models and more traditional machine learning models, pre-trained speech models such as Whisper, capture both semantic and non-semantic acoustic cues that are indicative of cognitive decline and have shown strong predictive performance even without manual transcription. Among the benchmarked speech and language models, Whisper-medium achieved the best performance on this large dataset. Our findings suggest that both semantic and non-semantic information are crucial for ADRD detection. Future acoustic-based ADRD detection tools should incorporate both semantic and non-semantic features. The use of acoustic-based automated ADRD detection tools offers a scalable, non-invasive, and cost-effective approach for population-level surveillance that supports earlier intervention and treatment. Technologies such as telehealth and mobile health further enhance their potential for real-world clinical deployment.

# Declarations

## Ethics Approval

This study was determined to be exempt from IRB.

## Data Availability

Publicly available datasets were analyzed in this study. This data can be found at: https://dementia.talkbank.org/DrivenData/

## Funding Information

The authors report no funding for this study.

## Conflict of Interest

The authors report no conflict of interest.

## Authors' Contributions

**JL, LM, XM, HW**, and **XSN** developed methodology and wrote the original draft. **JL and LM** curated data, conducted formal analysis, and prepared all the figures. **All authors** conceptualized the work, wrote, reviewed, edited, and approved final version of the draft.

## Acknowledgement

We want to thank DrivenData PREPARE Challenge for providing the data and making this study possible.

# Tables and Figures

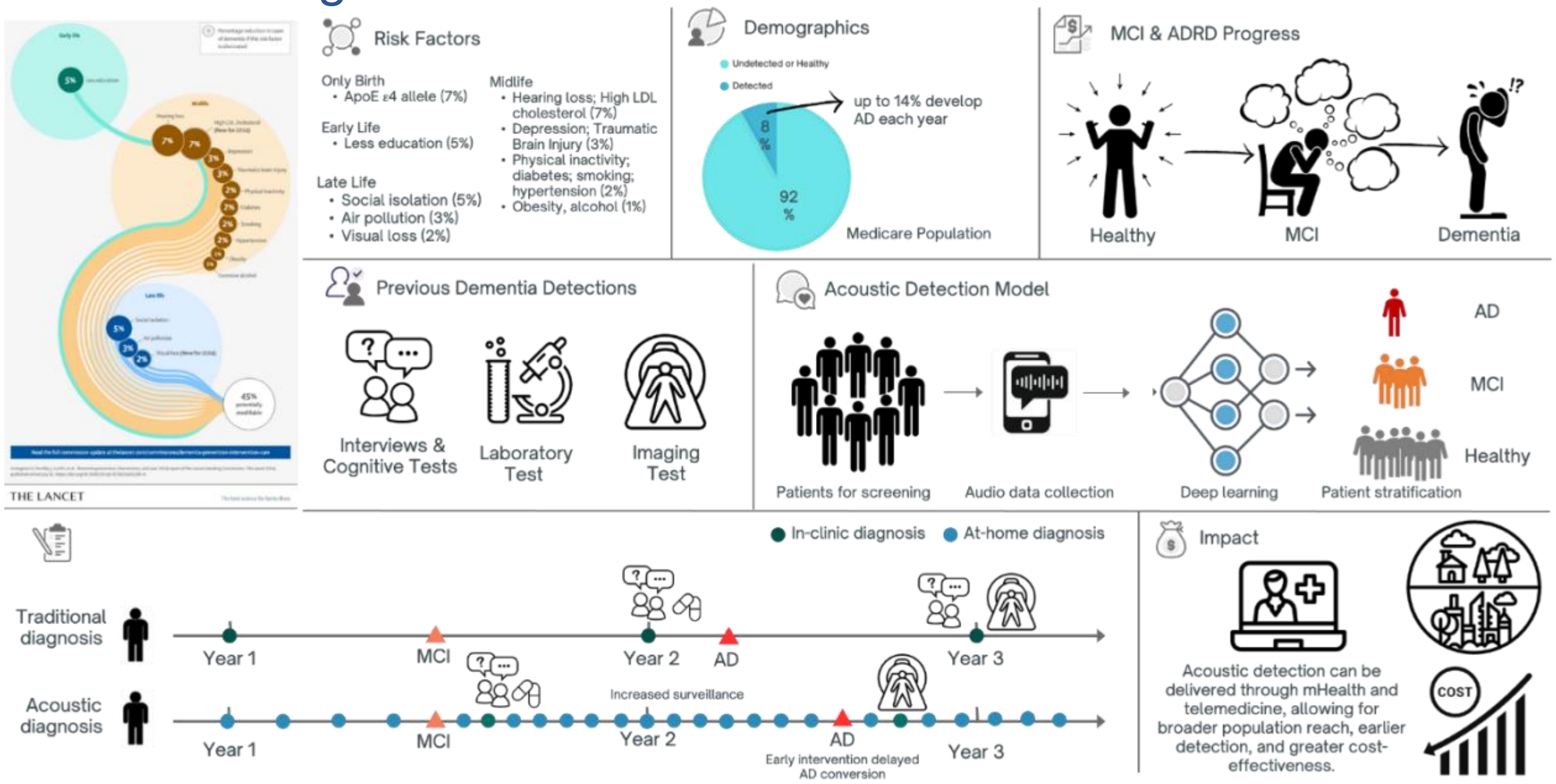

**Figure 1.** MCI and ADRD Progression and Detection. We illustrate the life-course risk factors for dementia, the progression from healthy aging to mild cognitive impairment (MCI) and Alzheimer's disease (AD), and the potential of acoustic detection as an innovative diagnostic tool. In the acoustic-based diagnosis framework, individuals provide spontaneous speech recordings via a mobile device or with clinician assistance. Pre-trained deep learning models extract acoustic embeddings from speech, which are then used to classify the individual's cognitive status as healthy, MCI, or AD. The model outputs can recommend an in-clinic follow-up for formal diagnosis and potential intervention. This framework enables convenient, flexible, and scalable surveillance for early detection of ADRD.

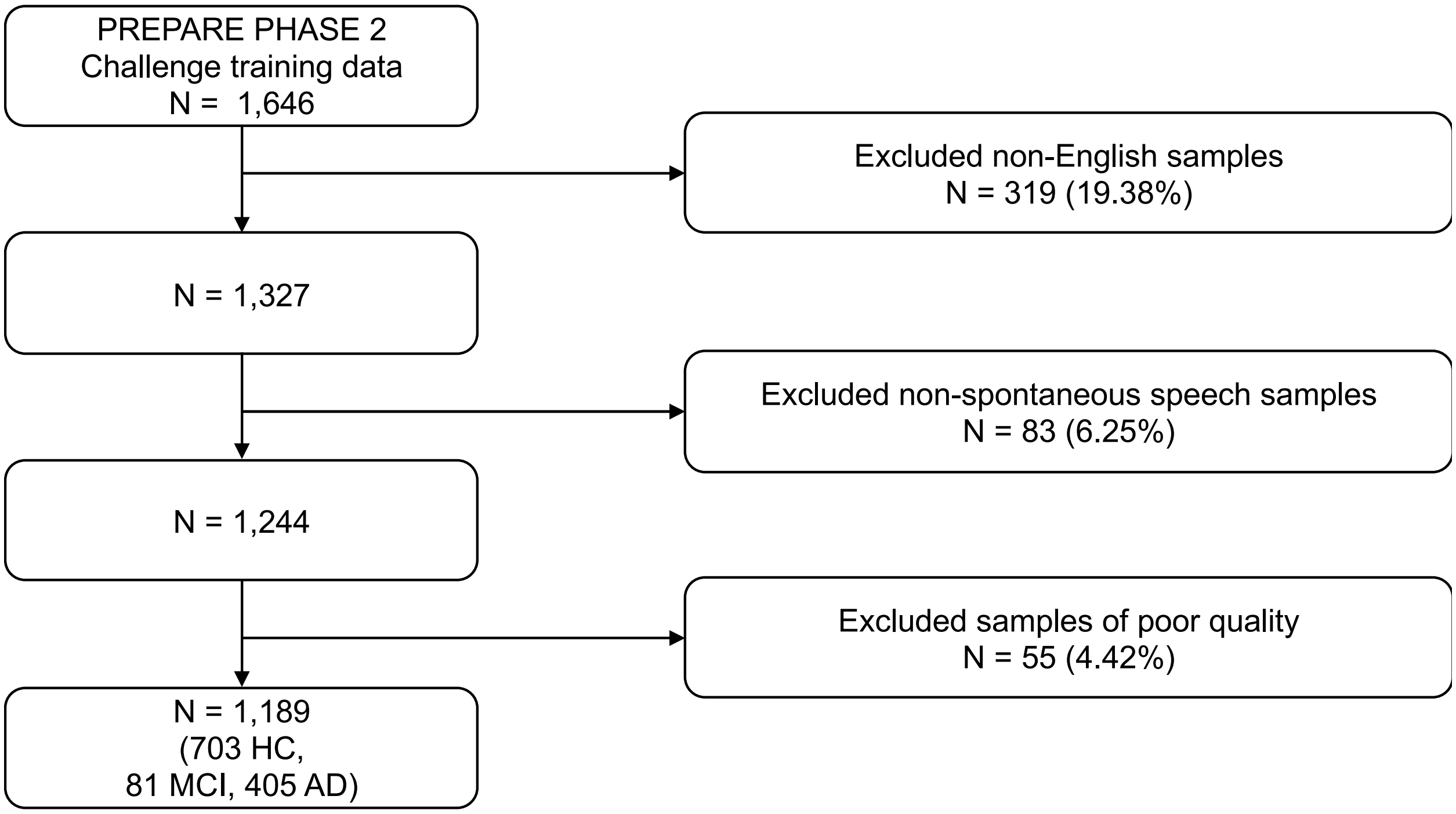

**Figure 2.** Data inclusion criteria. Speech samples were excluded if they were non-English, non-spontaneous, or of poor audio quality. The final dataset included 1,189 samples: 703 from healthy controls (HC), 405 from individuals with Alzheimer's disease (AD), and 81 from individuals with mild cognitive impairment (MCI).

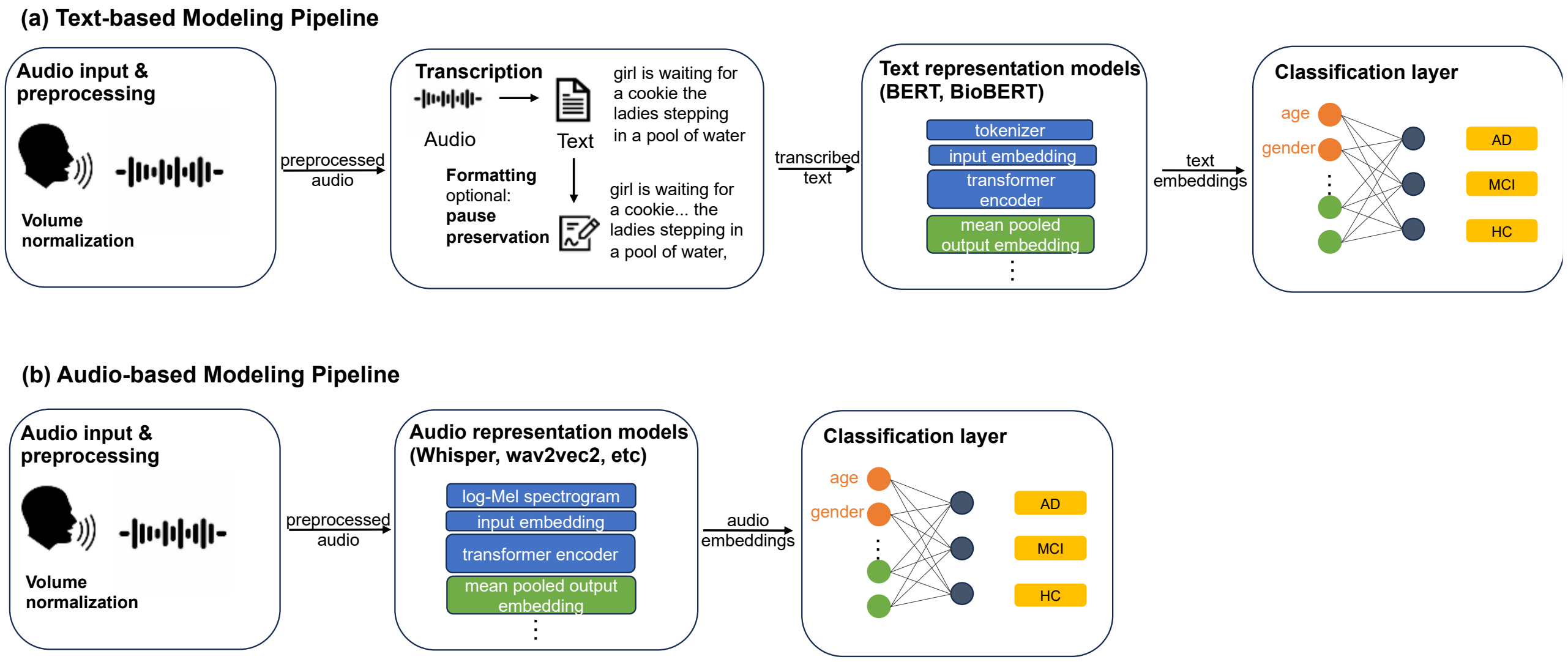

**Figure 3.** Overview of the modeling pipeline for ADRD detection from spontaneous speech samples. 2(a): text-based modeling pipeline: the system takes and preprocesses audio input, transcribes audio into text (optional: pause is annotated), extracts sentence-level embeddings from transcribed text using a text representation model, and feeds embeddings into a neural network to determine patient's cognitive status into three categories: healthy control (HC), mild cognitive impairment (MCI), or Alzheimer's disease (AD). 2(b): audio-based modeling pipeline: the system takes and preprocesses audio input, extracts sentence-level embeddings from audio data directly using an audio representation model, and feeds embeddings into a neural network to determine patient's cognitive status. The extracted embeddings are concatenated with demographic information before feeding into the classification layer.

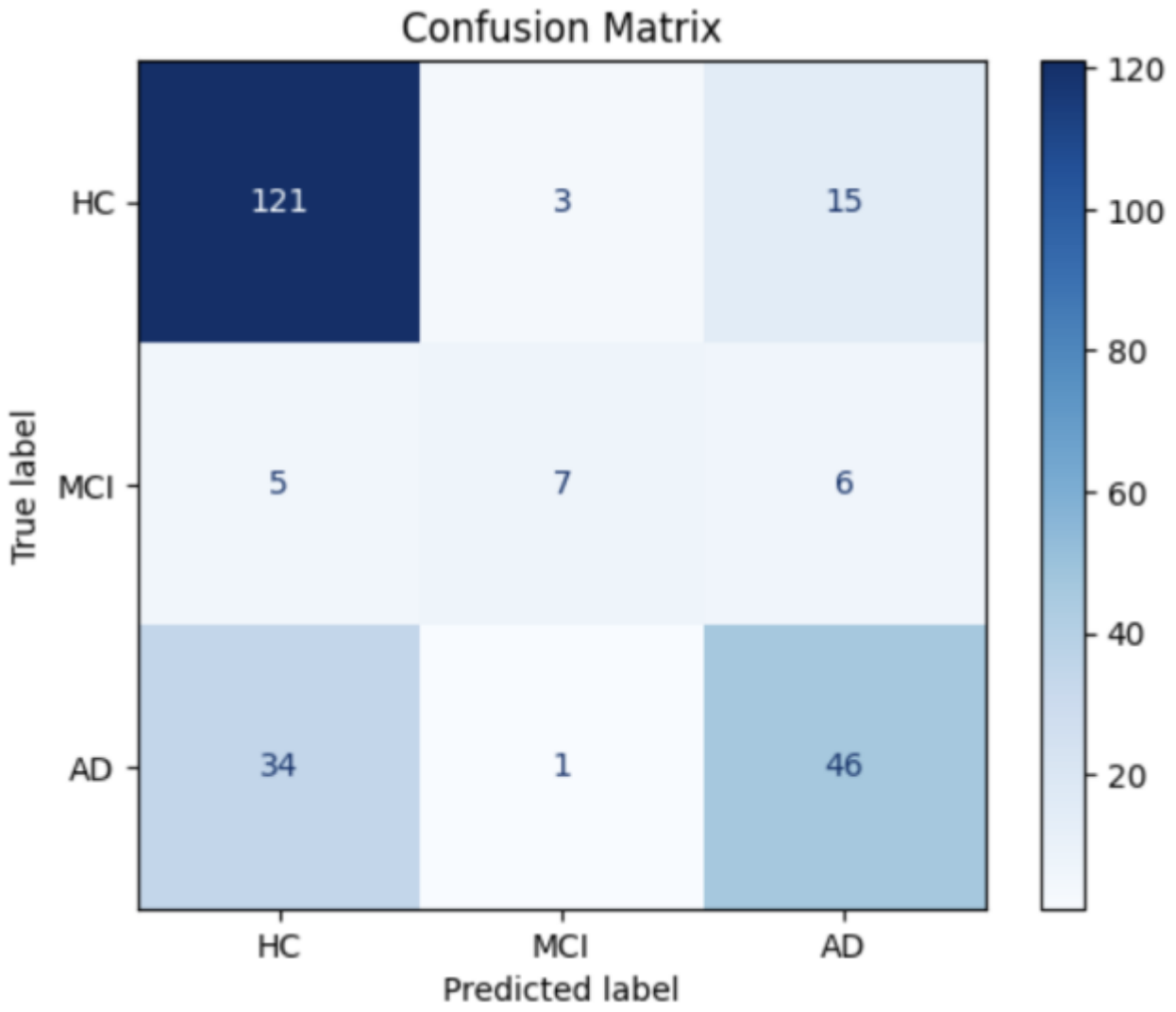

**Figure 4.** Confusion matrix of the best performing model (Whisper-medium)'s predictions on one replication of the test dataset. HC: healthy controls. MCI: mild cognitive impairment. AD: Alzheimer's disease.

**Table 2**. Test results for text-based models across 5 replications

| Models | Data size[a] | Model size | Accuracy | AUC[b] |
|---|---|---|---|---|
| Transcriptions by Whisper-small | | | | |
| BERT without pause | 3.3B | 110M | 0.6566 (0.0425) | 0.7186 (0.0257) |
| BERT with pause | 3.3B | 110M | **0.6622 (0.0131)** | 0.7258 (0.0158) |
| BioBERT without pause | 18B | 110M | 0.6122 (0.0329) | 0.7199 (0.0133) |
| BioBERT with pause | 18B | 110M | 0.6476 (0.0073) | 0.7286 (0.0112) |
| BioClinicalBERT without pause | 0.5B | 110M | 0.6258 (0.0394) | 0.7126 (0.0232) |
| BioClinicalBERT with pause | 0.5B | 110M | 0.6501 (0.0210) | 0.7118 (0.0221) |
| Transcriptions by Whisper-medium | | | | |
| BERT without pause | 3.3B | 110M | 0.6468 (0.0335) | 0.7207 (0.0083) |
| BERT with pause | 3.3B | 110M | 0.6539 (0.0284) | **0.7444 (0.0136)** |
| BioBERT without pause | 18B | 110M | 0.6287 (0.0209) | 0.7294 (0.0201) |
| BioBERT with pause | 18B | 110M | 0.6292 (0.0363) | 0.7375 (0.0335) |
| BioClinicalBERT without pause | 0.5B | 110M | 0.6233 (0.0244) | 0.7055 (0.0179) |
| BioClinicalBERT with pause | 0.5B | 110M | 0.6362 (0.0247) | 0.7198 (0.0313) |

a. Training data size in additional words trained. BioBERT were trained on additional words of 18B compared to BERT, and BioClinicalBERT were trained on additional words of 0.5B compared to BioBERT.
b. AUC: area under the curve.

**Table 3.** Test results for audio-based models across 5 replications

| Models | Data size[a] | Model size | Accuracy | AUC[b] |
|---|---|---|---|---|
| Whisper-tiny.en | 680,000h | 39M | 0.6733 (0.0391) | 0.7594 (0.0104) |
| Whisper-base.en | 680,000h | 74M | 0.7066 (0.0255) | 0.8006 (0.0200) |
| Whisper-small.en | 680,000h | 244M | 0.7011 (0.0328) | 0.7910 (0.0177) |
| Whisper-medium.en | 680,000h | 769M | **0.7307** (0.0202) | **0.8024** (0.0143) |
| Whisper-large | 680,000h | 1.55B | 0.7019 (0.0283) | 0.7787 (0.0137) |
| unispeech-large-1500h-cv | 1500h | 94M | 0.6589 (0.0230) | 0.7366 (0.0144) |
| wavlm-base | 960h | 94M | 0.6528 (0.0125) | 0.7548 (0.0323) |
| wavlm-base-plus | 94,000h | 94M | 0.6628 (0.0377) | 0.7477 (0.0177) |
| Wavlm-large | 94,000h | 316M | 0.6941 (0.0155) | 0.7761 (0.0231) |
| wav2vec2-base | 960h | 95M | 0.6215 (0.0275) | 0.7197 (0.0183) |
| wav2vec2-large | 960h | 137M | 0.6422 (0.0209) | 0.6938 (0.0295) |
| hubert-base | 960h | 95M | 0.6825 (0.0265) | 0.7517 (0.0308) |
| hubert-large | 960h | 317M | 0.6451 (0.0247) | 0.7254 (0.0114) |
| data2vec-audio-base | 960h | 94M | 0.5589 (0.0624) | 0.5317 (0.0248) |
| data2vec-audio-large | 960h | 317M | 0.6292 (0.0320) | 0.6667(0.0303) |

a. Training data size in hours of audio.

b. AUC: area under the curve.

**Table 4.** Test results for traditional feature engineering pipeline across 5 replications

| Feature | Model | Accuracy | AUC[a] |
|---|---|---|---|
| eGeMaps | Extra Trees | **0.6427** (0.0197) | **0.7669** (0.0128) |
| ComParE | Random Forest | 0.5878 (0.0211) | 0.6768 (0.0117) |
| MFCC | Random Forest | 0.6348 (0.0200) | 0.7538 (0.0130) |
| LIWC | Logistic Regression | 0.5834 (0.0129) | 0.6387 (0.0228) |

[a]AUC: area under the curve.

# Supplementary Material

## S 1. Clustering of BERT embeddings for topic modelling

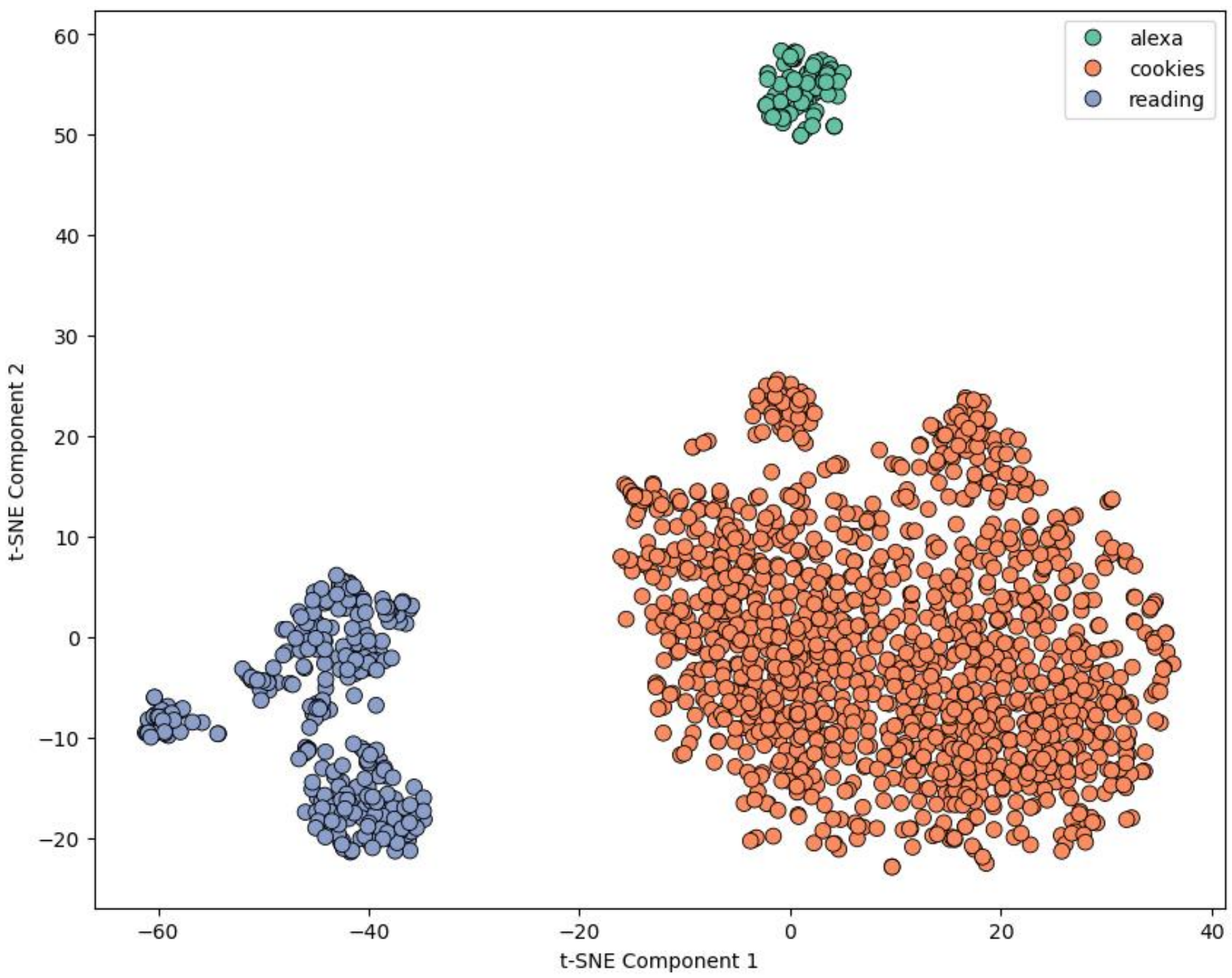

**Figure S1.** t-SNE visualization of clustering results on BERT embeddings of the original PREPARE Challenge dataset. Human evaluation classified the clusters into those performing spontaneous speech task (Cookies Theft Picture) and other sentence reading tasks.

## S 2. Traditional Feature Engineering Pipeline

**Audio preprocessing:** Each raw audio clip (48 kHz) is resampled to 16 kHz since high frequencies in the audio recordings typically do not contain speech. Audio amplitudes were normalized and beginning and end silence were trimmed. DeepFilterNet was used to remove background noise.

**Feature extraction:** We extracted multiple feature sets to capture acoustic and linguistic characteristics of speech. Acoustic features were computed using the openSMILE toolkit, including the extended Geneva Minimalistic Acoustic Parameter Set (eGeMAPS) [36], which captures prosodic, spectral, and voice quality features relevant to affective and clinical speech analysis, and the ComParE feature set [37], which provides a large-scale brute-force representation of low-level descriptors and functionals. In addition, 20-dimensional Mel-Frequency Cepstral Coefficients (MFCCs) [53] were computed to characterize the spectral envelope of the speech signal. For linguistic analysis, we used Linguistic Inquiry and Word Count (LIWC) to extract psychologically meaningful lexical features [39]. All features were standardized prior to modeling.

**Classification Models:** Each feature set was used independently to train binary classification models distinguishing Alzheimer's disease (AD) patients from cognitively normal controls. We employed PyCaret [54] to compare the model performance across multiple machine learning models including Extreme Gradient Boosting (XGBoost), Gradient Boosting Machine (GBM), Light GBM, Random Forest, Ada Boost Classifier, Ridge Classifier, Logistic Regression, Extra Trees, Decision Tree, K Neighbors Classifier, Linear Discriminant Analysis, Naive Bayes, and SVM. Hyperparameters were tuned via grid search using 5-fold cross-validation. The best performing model was chosen based on cross-validation AUC and the best model’s test performance across five independent replications was reported.

## S 3. Results of Whisper models with fine-tuning

**Table S1.** Test results for Whisper family of models across 5 replications with fine tuning

| Models | Accuracy | AUC[a] |
|---|---|---|
| Whisper-tiny.en | 0.6796 (0.0344) | 0.7612 (0.0172) |
| Whisper-base.en | 0.7091 (0.0257) | 0.7885 (0.0100) |
| Whisper-small.en | 0.7146 (0.0359) | 0.7999 (0.0250) |
| Whisper-medium.en | **0.7230** (0.0352) | **0.8019** (0.0278) |
| Whisper-large | 0.6979 (0.0187) | 0.7838 (0.0168) |

[a]AUC: area under the curve.
Last layers of Whisper models were fine-tuned while other layers were kept frozen during training.